\documentclass[letterpaper, 10 pt, conference]{ieeeconf}  % Comment this line out if you need a4paper

\IEEEoverridecommandlockouts                              % This command is only needed if 
\usepackage[utf8]{inputenc}
\usepackage{url}
\usepackage{graphics} % for pdf, bitmapped graphics files
\usepackage{graphicx} % for pdf, bitmapped graphics files
\usepackage{mathptmx} % assumes new font selection scheme installed
\usepackage{times} % assumes new font selection scheme installed
\usepackage{amsmath} % assumes amsmath package installed
\usepackage{amssymb}  % assumes amsmath package installed
\usepackage{color}
\usepackage{xcolor}
\usepackage{subfig}
\usepackage{epstopdf}
\usepackage{balance}
\usepackage{wrapfig}
\usepackage{adjustbox}
\usepackage{booktabs, makecell}
\usepackage{multirow}
\usepackage{times}
\usepackage{textcomp}
\usepackage{floatrow}
\usepackage{todonotes}
\usepackage{dsfont}
\usepackage{array}

\usepackage{algpseudocode}
\usepackage[ruled,linesnumbered]{algorithm2e}

\DeclareMathAlphabet{\mathcal}{OMS}{cmsy}{m}{n}

\newcommand{\norm}[1]{\left\lVert#1\right\rVert}
   
\renewcommand{\[}{\begin{equation}}
\renewcommand{\]}{\end{equation}}

\long\def\ignore#1{}

\usepackage{hyperref}

\usepackage{xcolor}
\usepackage{mdframed}

\newlength\myindent
\renewcommand{\vec}[1]{\mathbf{#1}}

\newcommand{\slim}{{\em SLiM}}

\title{\LARGE \bf
Semantic Linking Maps for Active Visual Object Search
}

\author{Zhen Zeng\authorrefmark{1} \hspace{0.5cm} Adrian Röfer\authorrefmark{2} \hspace{0.5cm} Odest Chadwicke Jenkins\authorrefmark{1}
\thanks{\authorrefmark{1}Z. Zeng, O.C. Jenkins are with the Department of Electrical Engineering and Computer Science, Robotics Institute, University of Michigan, USA}
\thanks{\authorrefmark{2}A. Röfer is with the Department of Computer Science, University of Bremen, Germany}
}

\begin{document}

\maketitle

\begin{abstract}
We aim for mobile robots to function in a variety of common human environments. Such robots need to be able to reason about the locations of previously unseen target objects. Landmark objects can help this reasoning by narrowing down the search space significantly. More specifically, we can exploit background knowledge about common spatial relations between landmark and target objects. For example, seeing a table and knowing that cups can often be found on tables aids the discovery of a cup. Such correlations can be expressed as distributions over possible pairing relationships of objects. In this paper, we propose an active visual object search strategy method through our introduction of the Semantic Linking Maps (\slim) model. \slim \ simultaneously maintains the belief over a target object's location as well as landmark objects' locations, while accounting for probabilistic inter-object spatial relations. Based on \slim, we describe a hybrid search strategy that selects the next best view pose for searching for the target object based on the maintained belief. We demonstrate the efficiency of our \slim-based search strategy through comparative experiments in simulated environments. We further demonstrate the real-world applicability of \slim-based search in scenarios with a Fetch mobile manipulation robot.
\end{abstract}

\thispagestyle{empty}
\pagestyle{empty}

%%%%%%%%%%%%%%%%%%%%%%%%%%%%%%%%%%%%%%%%%%%%%%%%%%%%%%%%%%%%%%%%%%%%%%%%%%%%%%%%
%\input{scope.tex}
\section{Introduction}
\begin{figure}[t!]
    \centering
    \includegraphics[width=1.0\textwidth]{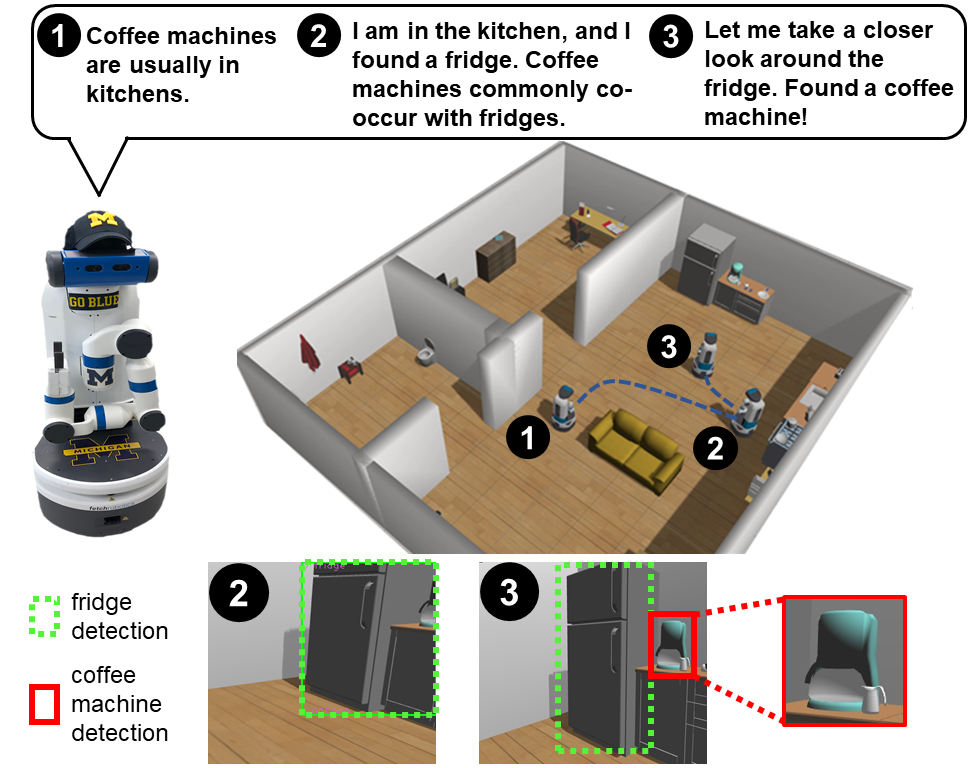}
    \caption{Robot tasked to find a coffee machine.}
    \label{fig:teaser}
\end{figure}

% general motivation for object search. novel target object. common spatial relations between novel target object and other objects. call them landmark objects. finding landmark objects are beneficial for object search tasks
Being able to efficiently search for objects in an environment is crucial for service robots to autonomously perform tasks~\cite{khandelwal2017bwibots,veloso2015cobots,hawes2017strands}. When asked where a target object can be found, humans are able to give hypothetical locations expressed by spatial relations with respect to other objects. For example, a \textit{cup} can be found ``on a table'' or ``near a sink". 
%We call these referenced objects 
\textit{Table} and \textit{sink} are considered landmark objects that are informative for searching for the target object \textit{cup}. Robots should be able to reason similarly about objects locations, as shown in Figure \ref{fig:teaser}.

Previous works~\cite{kollar2009utilizing, kunze2014using, toris2017temporal} assume landmark objects are static, in that they mostly remain where they were last observed. This assumption can be invalid for dynamic landmark objects that change their location over time, such as chairs, food carts and toolboxes. Temporal assumptions can mislead the search process 
%\rephrase{object process}{Do you mean search process?} 
if the prior on the landmarks' locations is too strong. %\ocj{provide an example where a strong temporal assumption causes an error}
Further, there also exists uncertainty in the spatial relations between landmark objects and the target object, and between landmark objects themselves. For example, a \textit{cup} can be ``in" or ``next to" a \textit{sink}.
%\ocj{provide an example of this uncertainty over spatial relations}

Considering the problem of dynamic landmarks, we propose the Semantic Linking Maps (\slim) model
%\rephrase{Semantic Linking Maps (\slim) model}{Verb and noun are not in agreement. Either \slim is singular, or \emph{models} is turned plural} 
to account for uncertainty in the locations of landmark objects during object search. 
% Sounds good! We're trying to get the system running on the actual robot right now.
% Great!  I am actually about to catch a flight.  Trying to get edits in while I can.
%Hi Adrian.  I am reorganizing the introduction to follow a format: problem/motivation then challenge wrt existing work then our approach and finishing with our contribution and synopsis of the paper  -chad
Building on Lorbach et al.~\cite{lorbach2014prior}, we model inter-object spatial relations probabilistically via a factor graph. The marginal belief on inter-object spatial relations inferred from the factor graph is used in \slim \ to account for probabilistic spatial relations between objects. %Some previous works~\cite{sjoo2012topological, aydemir2011search} assume known spatial relations between the target object and landmark objects in the environment (e.g. cup in box on table), then the robot can first purposefully look for those landmark objects first and then search for the target object in regions as specified by the given spatial relations. %This assumption is strong in that many target objects such as cups do not always keep the same spatial relations with other landmark objects. If the provided spatial relations are wrong, the robot is going to be misled. 
%In our work, we relax this assumption by learning and modeling inter-object spatial relations probabilistically. Thus we consider the probabilistic spatial dependencies between objects while maintaining the belief over the target and landmark objects locations.

Using the maintained belief over target and landmark objects' locations from \slim, we propose a hybrid strategy for active object search. We select the next best view pose, which guides the robot to explore promising regions that may contain the target and/or landmark objects. Previous works~\cite{wixson1994using, garvey1976perceptual, sjoo2012topological, aydemir2011search} have shown the benefit of purposefully looking for landmark objects (\textit{Indirect Search}) before directly looking for the target object (\textit{Direct Search}). The proposed hybrid search strategy draws insights from both indirect and direct search. We demonstrate the efficiency of the proposed hybrid search strategy in our experiments.%When the size of the landmark object is larger than the target object, it can be recognized more reliably from a longer distance. Once landmark objects are found, it helps narrow down the search region for the target object.

In this paper, we describe the Semantic Linking Maps model as a Conditional Random Field (CRF).  Our description of \slim~as a CRF allows us to simultaneously maintain the belief over target and landmark object locations with probabilistic modeling over inter-object spatial relations. We also describe a hybrid search strategy based on \slim~that draws upon ideas from both indirect and direct search representations.  This \slim-based search makes use of the maintained belief over objects' locations by selecting the next best view pose based on the current belief.
In our experiments, we show that the proposed object search approach is more robust to noisy priors on landmark locations by simultaneously maintaining 
%\rephrase{the}{That sounds odd. I think ith should be \emph{a belief}, since it is not specified what the belief is.} 
%\ocj{i might disagree. this use of belief is common in probablistic inference}
belief over the locations of target and landmark objects.

\section{Related Work}
% related works in object search: search in known environments; search in unknown environments
% in known environments, different kinds of prior is assumed.
% 1. assume prior knowledge on landmark objects
% different kind of dependence is being used: place/room-object, object-object, or both
% discuss if the prior is noisy, the performance degrades. Our experiments show the degration, and accounting for the uncertainty for the landmark objects location is important.
% 2. assume prior knowledge on spatial relationships between target and landmark objects
% strong prior; and if wrong, restrict object search to wrong region. we model the relations probabilistically instead.
Existing works have studied object search with different assumptions on prior knowledge of the environment. Some assume priors on landmark objects' locations in the environment, and utilize the spatial relations between the target object and landmark objects to prioritize regions to search. 
Kollar et al.~\cite{kollar2009utilizing} %pre-mapped the environment, including landmark objects, and 
utilize object-object co-occurrences extracted from image tags on Flickr.com to infer target object locations. Kunze et al.~\cite{kunze2014using} expanded the generic notion of co-occurrences to more restrictive spatial relations (e.g. ``in front of", ``left of"), which provide more confined regions to search, thus improving the search efficiency. %But the spatial relations between objects are assumed given rather than learned. 
Toris et al.~\cite{toris2017temporal} proposed to learn a temporal model on inter-object spatial relations to facilitate search. These methods assume the landmark objects to be static, however, we believe accounting for the uncertainty in landmark objects' locations is important for object search.

% Quote from Elfring et al.: "Among the popular alternatives are LabelMe [13], the Open Mind Common Sense (OMICS) [9] project and Flickr.com. In [15] LabelMe is used to get typical object-room relations, e.g., ‘kitchens typically contain a sink’ or ‘offices typically contain a bookshelf’. Afterwards, these statements are translated into probabil- ities. An approach leading to similar knowledge but using OMICS instead is presented in [11] whereas [10] uses Flickr for this purpose."

% prior on relation between landmark object and target object, indirect object search
Existing works have also explored known priors on spatial relations between landmark and target objects. Given exact spatial relations between landmark and target objects, Sj{\"o}{\"o} et al.~\cite{sjoo2012topological} used an \textit{indirect object search} strategy~\cite{wixson1994using, garvey1976perceptual}, where the robot first searches for landmark objects, and then searches for a target object in regions satisfying given spatial relations. On the other hand, given a probabilistic distribution over the spatial relations between objects, Aydemir et al.~\cite{aydemir2011search} formulate the object search problem as a Markov Decision Process. In our work, we learn the probabilistic inter-object spatial relations by building on ideas of Lorbach et al.~\cite{lorbach2014prior}, where inter-object relations are being probabilistically modeled via a factor graph.

% in unknown environments, 
% 1. rely on object-object, positive or negative detection on other landmark objects will result in updated belief over the target object. the relations being modeled is usually co-occurrences, which is usually mathematically being represented as a Gaussian around landmark objects. Instead, we represent more fine-grained spatial relations between objects, i.e., in, on, proximity, disjoint. the search process can be more specifically guided other than representing the distribution of the target object as a globe around the landmark object. In addition, "In" relation can trigger manipulation action to search for object by opening containers. Manipulation based object search is not within the scope of this paper, but the work can be used for manipulation based object search in future.
% 2. rely on object-place, place-place, Wang et al. Aydemir et al. does not account for object-object. More focused on path planning in large scale environment.

There are also works that do not assume prior knowledge of the environment. Researchers have explored object search with visual attention mechanisms~\cite{shubina2010visual, sjo2009object, meger2010curious}, such as saliency detection. Similar to~\cite{kollar2009utilizing, kunze2014using}, other research~\cite{loncomilla2018bayesian, elfring2013active, 5509285} utilizes object-object co-occurrences to guide the search for a target object. Positive and negative detections of landmark objects will result in an updated belief over the target object. % Joho et al.~\cite{5509285} designed vocabularies such as ``same aisle", ``same shelf" to express object-object co-occurrences in a supermarket setting. 
We expand object-object co-occurrences to finer-grained spatial relations between objects, i.e., ``in", ``on", ``proximity", ``disjoint", which specify more confined regions for object search.

Other literature~\cite{wang2018efficient, kunze2012searching, viswanathan2009automated} has also explored object-place relations to facilitate object search. Wang et al.~\cite{wang2018efficient} build a belief road map based on object-place co-occurrences for efficient path planning during object search. Kunze et al.~\cite{kunze2012searching} bootstraps commonsense knowledge on object-place co-occurrences from the Open Mind Indoor Common Sense (OMICS) dataset. Samadi learned similar knowledge by actively querying the World Wide Web (WWW). %Viswanathan et al.~\cite{viswanathan2009automated} use observed landmark objects to predict the type of places (e.g. \textit{kitchen}, \textit{office}) to inform the object search process, and they extract object-place co-occurrences from annotations of images in LabelMe~\cite{russell2008labelme}. In addition to object-place co-occurrences, 
Our work also takes object-place co-occurrences into account. Aydemir et al.~\cite{aydemir2013active} made use of place-place co-occurrences to infer the type of the room next door, as the robot explores an environment during search. Manipulation-based object search, as in~\cite{xiao2019online, wong2013manipulation, li2016act}, is not within the scope of this paper.

\section{Problem Statement}
%Let $O = \{o^1, o^2, \cdots, o^N\}$
Let $O = \{o^i | i = 1, \cdots, N\}$ be the set of objects of interest, including landmark objects and the target object for search. %Our work performs robotics search of target object informed by landmark objects as well as inter-object semantic spatial relation.
Given observations $z_{0:T}$ and robot poses $x_{0:T}$, we aim to maintain the belief over object locations $P(O_T|x_{0:T}, z_{0:T})$, while accounting for the probabilistic spatial relations $R_{ij}$ between objects $o^i, o^j \in O$. For this work, we consider the set of spatial relations  to be $R_{ij} \in \{\textit{In, On, Contain, Support, Proximity, Disjoint}\}$. For example, the relation $R_{ij}=In$  indicates that object $o_i$ is inside object $o_j$. The probabilistic spatial relations between object $o^i, o^j$ is represented by the belief over $R_{ij}$, denoted as $\mathcal{B}(R_{ij})$.

Based on the maintained belief $P(O_T|x_{0:T}, z_{0:T})$, the robot searches for the target object by selecting the next best view pose ranked by an utility function $U: \vec{\tau} \mapsto \mathbb{R}$. $\vec{\tau}$ specifies the 6 DOF of camera view pose. The utility function $U$ trades off between navigation cost and the probability of search success. Upon a user request to find a target object, the robot iterates between the belief update of objects' locations and view pose selection, until the target object is found or the maximum search time is reached.%We structure the proposed object search approach into: Sec. \ref{sec:maintain_object_belief} on \slim \ that maintains belief over object locations, while accounting for probabilistic inter-object spatial relations; Sec. \ref{sec:search_strategy} on the hybrid search strategy that selects the next best view pose by the proposed utility function.

% Each object $o^i=\{x, y, z\}$ contains the 3D location of that object. The spatial relation between a pair of object $(o^i, o^j)$ is denoted as $R_{ij}$.

% we maintain the belief over the location of landmark objects and the target object, while probabilistically modeling inter-object semantic spatial relations. And the robot searches for the target object by selecting the best next view pose ranked by an utility function, which trades off between navigation cost and the probability of search success.

%  The inter-object semantic spatial relation that we consider in this work is $R_{ij} \in \{$\textit{In, On, Contain, Support, Proximity, Disjoint}$\}$. For example, $R_{ij}=$\ \textit{In} indicates that object $o_i$ is inside object $o_j$. 

%Given observations $z_{0:T}$ of the environment and robot poses $x_{0:T}$, we maintain the belief over object locations $P(O_T|x_{0:T}, z_{0:T})$ while accounting for the probabilistic distribution of $R_{ij}$ between objects $o^i, o^j \in O$. 
%\input{methods.tex}
\section{Semantic Linking Maps}\label{sec:maintain_object_belief}
\begin{figure}[t!]
\centering
\includegraphics[width=0.85\textwidth]{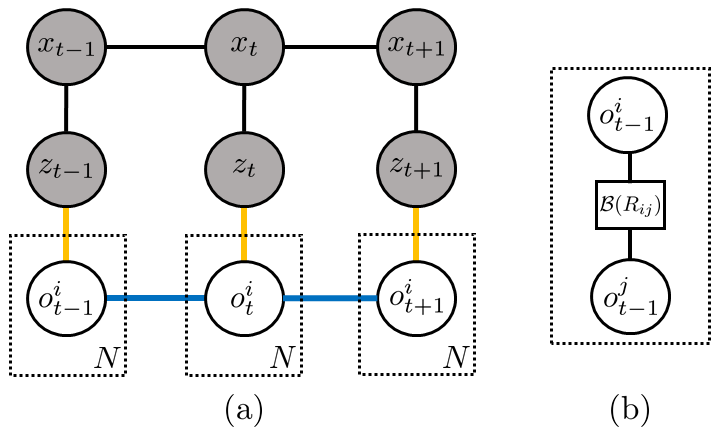}
\caption{CRF-based \slim \ model: (a) Known: $\{x^t\}$ robot poses, $\{z^t\}$ sensor observations; Unknown: $O_t = \{o^1_t, o^2_t, \cdots, o^N_t\}$. (b) Plate notation: at time $t$, the spatial relations between each object pair $o^i, o^j$ is parameterized by the belief over their spatial relations $\mathcal{B}(R_{ij})$.}
\label{fig:crf}
\end{figure}

For Semantic Linking Maps (\slim), we consider inter-object spatial relations, while maintaining the belief over target and landmark objects' locations. Building on our previous work~\cite{zeng2018semantic}, we probabilistically formalize the object location estimation problem via a Conditional Random Field (CRF). The model is now extended to account for probabilistic inter-object spatial relations, as shown in Figure \ref{fig:crf}.

The posterior probability of object locations $O$ history is
\begin{align}\label{eq:posterior_prob}
p(& O_{0:T} | x_{0:T},z_{0:T})= \nonumber \\
&\frac{1}{Z} \prod_{t=0}^T \prod_{i=1}^N \phi_p(o^i_t, o^i_{t-1}) \phi_m(o^i_t, x_t, z_t) \prod_{i,j} \phi_{c, \mathcal{B}(R_{ij})}(o_t^i, o_t^j)
\end{align}
where $Z$ is a normalization constant. Robot pose $x_{t}$ and observation $z_t$ are known. We assume that the robot stays localized given a metric map of the environment.%We explain three different potentials $\phi_p, \phi_m, \phi_c$ involved in CRF as below.

%\subsection{Prediction Potential}
$\phi_p(o^i_t, o^i_{t-1})$ is the \textit{prediction potential} that models the movement of an object over time. We assume objects to remain static or move with temporal coherence (varies across object classes) during the search, i.e.
\begin{equation*}
\phi_p(o^i_t, o^i_{t-1}) = e^{-(o^i_t-o^i_{t-1})^T \Sigma^{-1}(o^i_t-o^i_{t-1})}
\end{equation*}

%\subsection{Measurement Potential}\label{sec:measurement_potential}
$\phi_m(o^i_t, x_t, z_t)$ is the \textit{measurement potential} that accounts for the observation model, and $z_t=\{z^i_t | i=1, \cdots, N\}$ are (potentially noisy) detections for each object $o^i$ at time $t$. Because $z^i_t$ and $o^j$ are independent if $j \neq i$, we simplify $\phi_m(o^i_t, x_t, z_t)$ to $\phi_m(o^i_t, x_t, z^i_t)$ s.t.,
\begin{equation}\label{sec:measurement_potential}
\phi_m(o^i_t, x_t, z^i_t)=\begin{cases}
               P_{FN}, & \text{if $o^i_t \in E^i_t$ but $z^i_t=\emptyset$} \\
               P_{TN}, & \text{if $o^i_t \notin E^i_t$ and $z^i_t=\emptyset$} \\
               P_{TP}, & \text{if $\pi(o^i_t) \in z^i_t$} \\
               P_{FP}, & \text{otherwise}
            \end{cases}
\end{equation}
where each $P$ stands for the probability of false negative, true negative, true positive, and false positive detection. $E^i_t$ is the effective observation region for $o^i$ given robot pose at time $t$. Note, $E^i_t$ is larger for larger objects, which can be reliably detected from longer distance compared to small objects. $\pi$ is the camera projection matrix, and $\pi(o^i_t) \in z^i_t$ denotes that the projected object lies in the detected bounding box in $z^i_t$.

% \subsection{Context Potential}
We model the spatial relations between objects with \textit{context potential} $\phi_{c,\mathcal{B}(R_{ij})}$. Here, we extend $\phi_c$ from our previous work by parameterizing it with the belief $\mathcal{B}(R_{ij})$ over the inter-object spatial relation between $o^i, o^j$,
\begin{equation}\label{eq:context_potential}
\phi_{c,\mathcal{B}(R_{ij})} =  \sum\limits_{r} \mathcal{B}(R_{ij}=r) \phi_{c,r}(o^i_t, o^j_t, R_{ij}=r) 
\end{equation}
where $r$ can take any value in the set of possible relations $\{\textit{In, On, Contain, Support, Proximity, Disjoint}\}$. 

For $r \in \{\textit{In, On, Contain, Support}\}$, $\phi_{c,r}(o^i_t, o^j_t, R_{ij}=r)$ is equal to $1$ if objects $o^i_t, o^j_t$ satisfy the spatial relation given the width, length and height of the object, otherwise $0$. %Note that for $r = \text{In}$, we also consider the dependencies between objects and rooms. That is, we use auxiliary random variables $o^j$ representing rooms. And we assume a known area and category for each room.
For $r=\textit{Proximity}$, $\phi_{c,r}(o^i_t, o^j_t, R_{ij}=Proximity)$ corresponds to a Gaussian distribution that models $o^j_t \sim \mathcal{N}(o^i_t, \, \Sigma^{ij})$ and $\Sigma^{ij}$ is determined by the size of objects $o^i, o^j$. The larger the size of $o^i, o^j$, the larger the variance in $\Sigma^{ij}$.
For $r=\textit{Disjoint}$, $\phi_{c,r}(o^i_t, o^j_t, R_{ij}=Disjoint) = 1-\sum\limits_{r \neq Disjoint} \phi_{c,r}(o^i_t, o^j_t, R_{ij}=r)$.

% \begin{itemize}
%     \item For $r \in \{\textit{In, On, Contain, Support}\}$, $\phi_{c,r}(o^i_t, o^j_t, R_{ij}=r)$ is equal to $1$ if objects $o^i_t, o^j_t$ satisfies the spatial relation given the width, length and height of the object size, otherwise $0$. Note that for $r = \text{In}$, we also consider the dependencies between objects and rooms. That is, we use auxiliary random variables $o^j$ that represents rooms. And we assume known area and category for each room.
%     \item For $r=\textit{Proximity}$, $\phi_{c,r}(o^i_t, o^j_t, R_{ij}=Proximity)$ corresponds to a Gaussian distribution that models $o^j_t \sim \mathcal{N}(o^i_t, \, \Sigma^{ij})$ and $\Sigma^{ij}$ is determined on size of objects $o^i, o^j$. The larger the size of $o^i, o^j$, the larger the variance is in $\Sigma^{ij}$.
%     \item For $r=\textit{Disjoint}$, $\phi_{c,r}(o^i_t, o^j_t, R_{ij}=Disjoint) = 1-\sum\limits_{r \neq Disjoint} \phi_{c,r}(o^i_t, o^j_t, R_{ij}=r)$.
% \end{itemize}
 
\subsection{Inference}
\begin{figure*}[t!]
\centering
\includegraphics[width=0.95\textwidth]{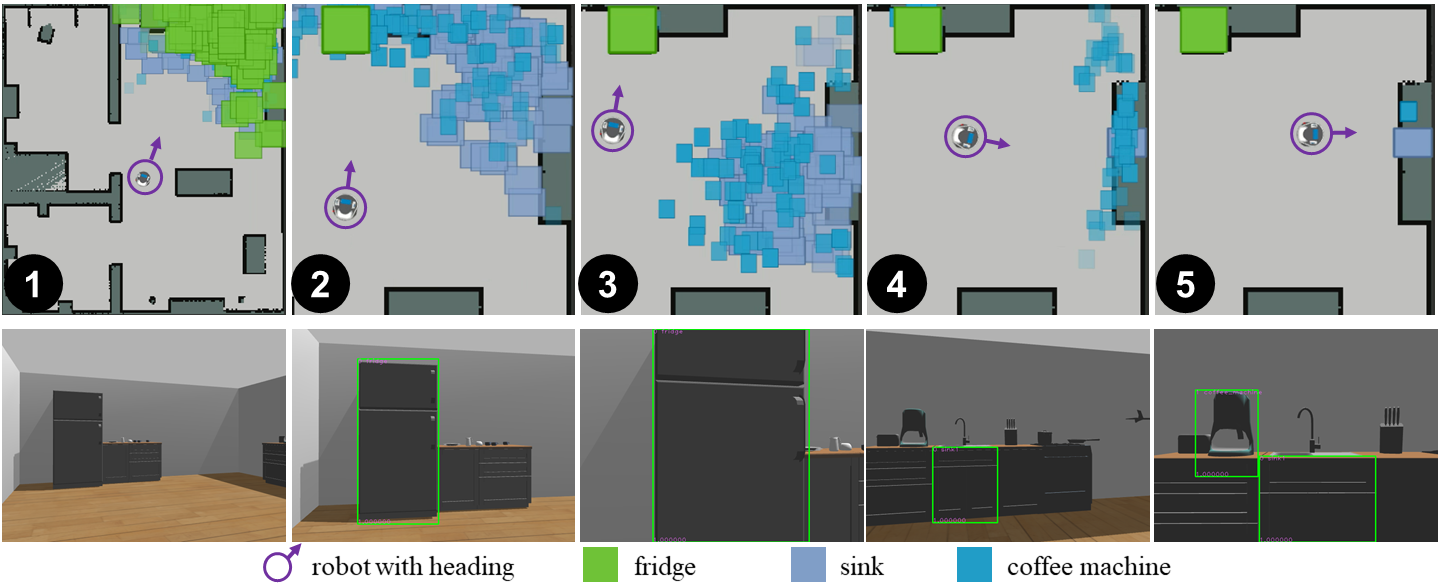}
\caption{Examples of belief updates in \slim. given observations. \textit{Upper}: Evolution of particles of \textit{fridge, sink, coffee machine} over time. \textit{Lower}: RGB observation (with object detection) over time. (Best viewed in color).}
\label{fig:belief_update}
\end{figure*}

We propose a particle filtering inference method for maintaining the belief over object locations, as shown in Algorithm \ref{alg:particle_filter}. Examples of the belief update over time are available in Figure \ref{fig:belief_update}. Instead of estimating the posterior of the complete history of object locations $p(O_{0:T} | x_{0:T},z_{0:T})$, we recursively estimate the posterior probability of each object $o^i_t \in O_t$, similarly to~\cite{zeng2018semantic, limketkai2007crf}. % i.e., $o_t^i=\{\langle o^{i(k)}_t, \alpha^{i(k)}_t \rangle | k=1, \cdots, M \}$, where $\alpha^{i(k)}_t$ is the associated weight for the $k^{th}$ particle. In each particle filtering iteration, particles are first resampled based on their associated weights, then propagated forward based on the prediction potential, and re-weighted according to the measurement and context potentials. 

%We represent each object with $M$ weighted particles. 
To deal with particle decay, we reinvigorate the particles of each $o^i$ by sampling in known room areas, as well as around other objects $o^j$ based on $\mathcal{B}(R_{ij})$. In step 5, $j \in \Gamma(i)$ only if $1-\mathcal{B}(R_{ij}=\textit{Disjoint}) > 0.2$. Across our experiments, we use 100 particles for each object. The inference algorithm does not assume single object instance for each object class. The inference algorithm has a complexity of $\mathcal{O}(nKM^2)$, where $K$ is the average cardinality of $\Gamma(i)$. Further works can be done to decrease the complexity down to $\mathcal{O}(nKMC)$ by sampling $C$ representative and divergent particles from the original $M$ particles ($C < M$).

%We only establish the edge of \textit{context potential} between objects $(o^i, o^j)$ if $1-\mathcal{B}(R_{ij}=\textit{Disjoint}) > 0.2$. Thus, for each object $o^i$, not necessarily all other objects are in the neighborhood $\Gamma(i)$ (step 5). 

% \begin{algorithm}[t!]
% \KwIn{Observation $z_t$, robot pose $x_t$, particle set for each object $Q_{t-1}^i=\{\langle o^{i(k)}_{t-1}, \alpha^{i(k)}_{t-1} \rangle | k=1, \cdots, M \}$}
% Resample $M$ particles $o^{i(k)}_{t-1}$ from $Q_{t-1}^i$ with probability proportional to importance weights $\alpha^{i(k)}_{t-1}$ \;
% \For{$i=1, \cdots, N$}{
% 	\For{$k=1, \cdots, M$}{
% 		Sample $o^{i(k)}_t \sim \phi_p(o^{i}_t, o^{i(k)}_{t-1}, u_{t-1})$ \;
% 		Assign weight $\alpha^{i(k)}_t \propto \phi_m(o^{i(k)}_t, x_t, z_t) \prod_{j \in \Gamma(i)} \phi_c(o^{i(k)}_t, o^j_{t-1})$ \;
% 	}
% }
%  \caption{Particle filtering in {\em CT-Map}} \label{alg:particle_filter}
% \end{algorithm}
\begin{algorithm}[t!]
\KwIn{{\begin{minipage}[t]{0.7\textwidth}%
        \strut
        Observation $z_t$, Robot pose $x_t$, \\
        Particle set for each object:  \\
        $\text{\ \ \ \ \ \ \ \ }o_{t-1}^i=\{\langle o^{i(k)}_{t-1}, \alpha^{i(k)}_{t-1} \rangle | k=1, \cdots, M \}, i \in 1:N$
        \strut
   \end{minipage}%
}}
Resample $M$ particles $o^{i(k)}_{t-1}$ from $o_{t-1}^i$ with probability proportional to importance weights $\alpha^{i(k)}_{t-1}$ \;
\For{$i=1, \cdots, n$}{
	\For{$k=1, \cdots, M$}{
		Sample $o^{i(k)}_t \sim \phi_p(o^{i}_t, o^{i(k)}_{t-1})$ \;
		Assign weight $\alpha^{i(k)}_t \propto \phi_m(o^{i(k)}_t, x_t, z_t) \prod\limits_{j \in \Gamma(i)} \phi_{c,\mathcal{B}(R_{ij})}(o^{i(k)}_t, o^j_{t-1})$ \;
		\ \ \ \  where $\phi_{c,\mathcal{B}(R_{ij})}(o^{i(k)}_t, o^j_{t-1}) =$ \
		$\sum\limits_{r} \sum\limits_{l=1}^M \mathcal{B}(R_{ij}=r) \alpha_{t-1}^{j(l)} \phi_{c,r}(o^i_t, o^j_t, R_{ij}=r) $
% 		\sum\limits_{l=1}^M \sum\limits_{\varepsilon=1}^N \phi_{c,g}(g_{t-1}^\varepsilon, o_t^{i(k)}, o_{t-1}^{j(l)}) \times \alpha_{t-1}^{j(l)} \times \alpha_{t-1}^\varepsilon$
	}
}
 \caption{Inference of objects locations in \slim.} \label{alg:particle_filter}
\end{algorithm}

\subsection{Probabilistic Inter-Object Spatial Relations}\label{sec:factor_graph}

To get the belief over inter-object spatial relations $\mathcal{B}(R_{ij})$ for each object pair $o^i, o^j \in O$, we use a factor graph by building on preceding work by Lorbach et al~\cite{lorbach2014prior}. We generalize~\cite{lorbach2014prior} by relaxing the assumption on known spatial relations between landmark objects.

%$F=\{F_{ij}, F_{ijk} | \forall_{i \neq j \neq k} \ o^i, o^j, o^k \in O\}$. 

The factor graph $G:\{\mathbb{V}, \mathbb{F}, \mathbb{E}\}$ consists of variable vertices $\mathbb{V}=\{R_{ij} | \forall_{i \neq j} \ o^i, o^j \in O\}$, factor vertices $\mathbb{F}=\{F_{CS}, F_{LC}\}$ and edges $\mathbb{E}$ which connect factor vertices with variable vertices. Specifically, $F_{CS}: R_{ij} \mapsto \mathbb{R}$ is a unary factor that considers \textit{commonsense knowledge} on spatial relation between objects,
\begin{equation*}
F_{CS}(R_{ij}) = \text{Frequency}(R_{ij})
\end{equation*}
Similar to ~\cite{lorbach2014prior}, we extract commonsense knowledge on $R_{ij}$ from online image search engine (e.g. Flickr) by counting the frequency of certain spatial relation between objects $o^i, o^j$. For example, the frequency of $R_{cup, table}=On$ is computed as the number of search results of a query ``cup on the table" divided by the number of search results of a query ``on the table". These extracted frequencies can be noisy. For example, the frequency of ``laptop on kitchen" is larger than 0, but it is not a valid expression because it refers to a laptop being on top of the room geometry of a kitchen. We manually encode the $F_{CS}(R_{ij})$ for invalid expressions to $0$.

$F_{LC}: (R_{ij}, R_{ik}, R_{jk}) \mapsto \{0, 1\}$ is a triplet factor that considers \textit{logical consistency} between a triplet of objects $o^i, o^j, o^k$,
\begin{equation*}
    F_{LC}(R_{ij}, R_{ik}, R_{jk})=\begin{cases}
    1, & \text{if consistent}.\\
    0, & \text{otherwise}.
  \end{cases}
\end{equation*}
For example, if $o^i$ is in $o^j$, and $o^j$ is in $o^k$, then $o^i$ should be in $o^k$ to satisfy logical consistency, i.e., $F_{LC}(R_{ij}=\textit{In}, R_{ik}=\textit{In}, R_{jk}=\textit{In}) = 1$. %n the other hand, if $o^i$ is in $o^j$, but $o^j$ is disjoint with $o^k$, then $o^i$ should also be disjoint with $o^k$ to satisfy logical consistency, i.e. $F_{LC}(R_{ij}=\text{In}, R_{ik}=\text{Disjoint}, R_{jk}=\text{Disjoint}) = 1$. 
%Previous work~\cite{lorbach2014prior} assumes known spatial relations between landmark objects, and the only unknown spatial relations are $R_{target, j}$ between the target object $o_{target}$ and the landmark objects $o_j$. Thus they only designed a pairwise factor for \textit{logical consistency}, i.e., $F_{LC}: (R_{target,j}, R_{target, k}) \mapsto \{0, 1\}$. In contrast, we design a triplet factor $F_{LC}$ that considers all possible combinations of $(R_{ij}, R_{ik}, R_{jk})$ and evaluate whether it is logically consistent.
Previous work~\cite{lorbach2014prior} assumes the spatial relations between landmark objects to be known, and only relations $R_{target, j}$ connecting target object $o_{target}$ and landmark object $o_j$ to be unknown. Their pairwise factor enforcing \textit{logical consistency} is a binary function $F_{LC}: (R_{target,j}, R_{target, k}) \mapsto \{0, 1\}$. In contrast, our formulation employs a trinary factor $F_{LC}$ considering all possible combinations of $(R_{ij}, R_{ik}, R_{jk})$ and evaluating their logical consistency.

By applying Belief Propagation~\cite{kschischang2001factor} on the factor graph formulated as above, we can get the marginal belief over inter-object relations $\mathcal{B}(R_{ij})$ between all object pairs. We use the libDAI~\cite{mooij2010libdai} library for inference. An example of the probabilistic inter-object spatial relations inferred from the factor graph is as shown in Figure \ref{fig:relation_marginal}, and it is used in our experiments.

% \begin{figure}[t!]
%     \centering
%     \includegraphics[width=0.9\textwidth]{figures/factor_graph.png}
%     \caption{We use a factor graph to model the semantic relation $R_{ij}$ between any two objects $o^i, o^j$: \textbf{CS}: common sense knowledge, \textbf{LC}: logical consistency.}
%     \label{fig:factor_graph}
% \end{figure}

% \begin{algorithm}[t!]
% \small
% \caption{Build Scene Graph from Factor Graph}\label{factor2scene}
% \begin{algorithmic}[1]
% \State \textbf{Input:} $target\ object\ o_i,
% % \ marginal\ belief (r_{ij})\  P_r(o_i, o_j),
% \ neighbor\ threshold\ \delta \newline
% marginal\ belief\ from\ factor\ graph\ P_r(o_i,\,o_j)$

% \Function{ExtractSceneGraph}{$o_i, \;P_r(o_i,\,o_j), \;\delta$}
% \If{$type(o_i)\ ==\ SUPPORTER$}
%   \State \Return
% \ElsIf{$type(o_i)\ ==\ CONTAINER$}
% %   \State $n.parent\gets MostLikelyParent(FG,o_i)$
%   \State $n.parent\gets \argmax_{o_j}P_{on}(o_i,\,o_j)$
% %   \State $n.neighbors\gets LikelyNeighbors(FG,o_i,\delta)$
%   \State $n.neighbors\gets \{\forall_{o_j}\,:\,P_{proximity}(o_i,\,o_j)>\delta\}$
% \Else
%   \State $n.parent\gets \argmax_{o_j}P_{on|in}(o_i,\,o_j)$
%   \State $n.neighbors\gets \{\forall_{o_j}\,:\,P_{proximity}(o_i,\,o_j)>\delta\}$
%   \State $ExtractSceneGraph(n.parent, \;P_r(o_i,\,o_j), \;\delta)$
% \EndIf
% \EndFunction

% \end{algorithmic}
% \end{algorithm}
\section{Search Strategy}\label{sec:search_strategy}
Based on the belief over the object locations, we actively search for the target object, by generating promising view poses and select the best one ranked by a utility function. %The view pose with the highest utility score is selected as the robot's next goal. Once the robot reaches the selected view pose, the belief over the object locations gets updated as the new observations become available.
Given the particle set $\langle o^{(k)}_t, \alpha^{(k)}_t \rangle$ of the target object $o$ as being maintained in \ref{sec:maintain_object_belief}, we fit Gaussian Mixture Models (GMMs) through Expectation Maximization to the particles by auto selecting the number of clusters~\cite{figueiredo2002unsupervised},
\begin{equation}\label{eq:gmms}
\langle o^{(k)}_t, \alpha^{(k)}_t \rangle \sim \langle \mathcal{N}(x_n, \Sigma_n), \omega_n \rangle
\end{equation}

\subsection{View Pose Generation}
For each Gaussian component $\mathcal{N}(\vec{x}_n, \Sigma_n)$, we generate a set of camera view pose candidates $\{\vec{\tau}_n^i = (\vec{c}_n^i, \vec{\psi}_n^i)\}$, where $\vec{c}_n$ and $\vec{\psi}_n$ denote the translation and the rotation of the camera respectively.

Initially, we sample the location of the camera $\vec{c}_n$ evenly from a circle with a fixed radius around the center $\vec{x}_n$ of the Gaussian component, and assign a default value to rotation $\vec{\psi}_n$. Note, that these initially sampled view poses can put the robot in collision with the environment, and the camera is not necessarily looking at $\vec{x}_n$. Thus, we formulate a view pose optimization problem under constraints as below,%\ocj{what expression shoud be on the left hand side of this equation?}
\[
\label{eq:opt_viewpose}
\begin{aligned}
\underset{\vec{\tau}_n}{\text{argmin}}\ 1 - \vec{v}_n \cdot \frac{\vec{x}_n - \vec{c}_n}{\norm{\vec{x}_n - \vec{c}_n}}\hspace{0.2cm}%\\ 
  \text{s.t} \
  & \vec{x}_n \in E_{\vec{\tau}_n}, 
  & c(\vec{\tau}_n) > 0
\end{aligned}
\]
where $\vec{v}_n$ is the view direction given $\vec{\tau}_n$, $E_{\vec{\tau}_n}$ denotes the effective observation region of the target object at camera pose $\vec{\tau}_n$, and $c: \vec{\tau} \mapsto \mathbb{R}$ is a function that computes a signed distance of a configuration $\vec{\tau}$ to the collision geometry of the environment.

\subsection{View Pose Selection}
We propose two different utility functions to rank the view pose candidates:
% \begin{itemize}
%     \item \textbf{Direct Search} utility $\mathbf{U}_{\text{DS}}$
%     \item  \textbf{Hybrid Search} utility $\mathbf{U}_{\text{HS}}$
% \end{itemize}
\subsubsection{\textbf{Direct Search} utility}
$\mathbf{U}_{\text{DS}}$ encourages the robot to explore promising areas that could contain the target object while accounting for navigation cost,
\begin{equation}\label{eq:ds}
    \mathbf{U}_{\text{DS}}(\vec{\tau}_k) = \omega_n + \alpha \frac{1}{\arctan(\sigma d_{nav})}
\end{equation}
where $\omega_n$ is the weight of the Gaussian component (as in (\ref{eq:gmms})) that $\vec{\tau}_k$ is generated from, and $d_{nav}$ is the navigation distance from the current robot location to view pose $\vec{\tau}_k$. Parameter $\alpha$ trades off between the probability of finding the target object and the navigation cost. Parameter $\sigma$ determines how quickly the $\arctan(\sigma d_{nav})$ plateaus.

With $\mathbf{U}_{\text{DS}}$, the object search is \textbf{direct} because we are directly considering promising areas represented by the GMMs for the target object.

% With $\mathbf{U}_{\text{DS}}$, the object search is \textbf{informed} because the belief over the target objects are informed by the belief over landmark objects in the environment, and the object search id \textbf{direct} because we are directly considering promising areas represented by the GMMs for the target object.

% \begin{figure*}[t!]
%     \centering
%     \includegraphics[width=1.0\textwidth]{figures/probabilities.png}
%     \caption{Marginal beliefs on inter-object spatial relations, as well as object-room relation, inferred from the factor graph as explained in Sec. \ref{sec:factor_graph}.}
%     \label{fig:relation_marginal}
% \end{figure*}

\begin{figure}[t!]
\centering
\includegraphics[width=0.7\textwidth]{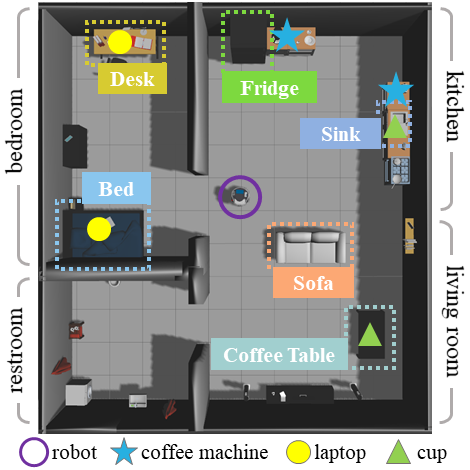}
% \includegraphics[width=0.2994\textwidth]{figures/gazebo_annotation_fit.png}
% \includegraphics[width=0.274\textwidth]{figures/gazebo_targets_fit.png}
%\includegraphics[width=0.3\textwidth]{figures/gazebo2.png}
% \caption{\textit{Left}: Gazebo simulation in an apartment-like environment with four rooms. \textit{Middle}: $6$ landmark objects used in experiments. \textit{Right}: Possible locations of $3$ target objects: coffee machine, laptop, cup.}
\caption{Simulation experiments setup in Gazebo: an apartment-like environment with four rooms. There are $6$ landmark objects and $3$ target objects: \textit{coffee machine, laptop, cup}. Each target object has two equally possible locations.}
\label{fig:env_setup}
\end{figure}

\subsubsection{\textbf{Hybrid Search} utility}
$\mathbf{U}_{\text{HS}}$ encourages the robot to explore promising areas that could contain the target object and/or any landmark object, while accounting for navigation cost
\[
\label{eq:hs}
\begin{aligned}
\mathbf{U}_{\text{HS}}(\vec{\tau}_k) = \ 
& \omega_n + \alpha \frac{1}{\arctan(\sigma d_{nav})} \\
& + \beta \ \max_{j,n} \ \text{CoOccur}(o, o^j) \omega_n^j \mathbf{I}^j_n
\end{aligned}
\]
where the additional term compared to $\mathbf{U}_{\text{DS}}$ acts to encourage the robot to also explore areas that could contain landmark object $o^j$ which co-occurs with the target object $o$ with probability $\text{CoOccur}(o, o^j)$. Specifically, $\text{CoOccur}(o, o^j) = (1-\mathcal{B}(R_{target, j}=\textit{Disjoint}))$, and $\omega^j_n$ is the weight of the $n$-th Gaussian component of GMMs fitted to the belief over the location of the landmark object $o^j$. And $\mathbf{I}^j_n$ is $1$ if the $n$-th Gaussian of object $o^j$ is within the effective observation region at camera pose $\vec{\tau}_k$, otherwise $0$.

$\mathbf{U}_{\text{HS}}$ is inspired by the \textit{indirect object search} strategy as studied in~\cite{garvey1976perceptual, wixson1994using}. Previous studies demonstrated that purposefully looking for an intermediate landmark object helps quickly narrow down the search region for the target object if the landmark object often co-occurs with the target object, thus improving the search efficiency.

With $\mathbf{U}_{\text{HS}}$, the object search can be considered \textbf{hybrid} because we are considering promising areas represented by GMMs for both the target object (as in direct search) and landmark objects that co-occur with the target object (as in indirect search).

In our experiments, we use a A$^\ast$ based planner to compute $d_{nav}$. We empirically set $\alpha=0.1$, $\beta=0.4$ , and $\sigma=0.5$ such that $\arctan(\sigma d_{nav})$ plateaus as $d_{nav}$ goes beyond 3$m$.
\section{Experiments}
% simulation experiments, real experiment
We perform object search tasks in both simulation and real-world environments with a Fetch robot. In the simulation experiments, we quantitatively benchmark various methods, including methods that resemble previous works and our proposed method. In the real-world experiments, we demonstrate qualitatively that the proposed method scales to real-world applications. In both simulation and real-world experiments, the robot accelerates to at most $1$m/s and turns at most at $1.7$rad/s.

\subsubsection{Simulation Experiments}
\begin{figure}[t]
    \centering
    %\hspace*{-0.2cm}
    \includegraphics[width=1\columnwidth]{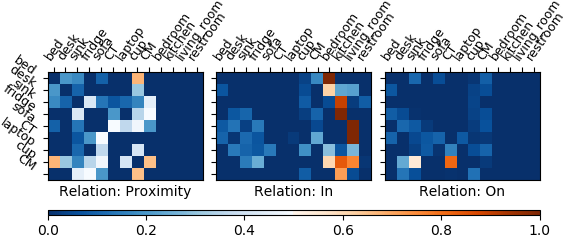}
    \caption{Marginal belief on inter-object spatial relations, as well as object-room relations, inferred from the factor graph as explained in Sec. \ref{sec:factor_graph}.   \emph{CM}: coffee machine, \emph{CT}: coffee table}
    \label{fig:relation_marginal}
\end{figure}

% simulation experiment # of objects, size of map, placement of object, figure to show the set up of the environment
The simulation experiments are performed in an apartment-like environment ($10$mx$11$m) setup in the Gazebo simulator, as shown in Figure \ref{fig:env_setup}. The room types and considered landmark objects are annotated in Figure \ref{fig:env_setup}, along with the placements of target objects. %The target objects for the searches are: a coffee machine, a laptop and a cup. %We model the belief over the locations of the target objects as well as several landmark objects including sofa, coffee table, sink, fridge, bed and desk, while accounting for the probabilistic inter-object semantic spatial relations between all object pairs. 
The marginal belief $R_{ij}$ inferred from the factor graph as explained in \ref{sec:factor_graph} is depicted in Figure \ref{fig:relation_marginal}.

% "laptop", "bed", "sofa", "livingroom_table", "desk", "cup", "sink1", "fridge",  "coffee_machine", "livingroom", "bedroom", "kitchen", "restroom"

% semantic observer in simulation: sensor field of view, horizontal range, vertical range, min and max depth of ... (cutting plane); detection: fired up criteria; detection FPS in sim and real

%we assume an effective observation range of $5$m. For mid-sized objects (e.g. desk, table, sink), we assume an effective observation range of $4$m. For small objects (e.g. cup, laptop, coffee machine), we assume an effective observation range of $2.5$m to $3$m.

%(1) the ray-casting from the robot camera to the center of the object is not blocked by other object, (2) the 2D projected mask of the object in the image plane exceeds certain number of pixels, and (3) the distance between the center of the object and robot camera is within the effective observing range for that object. 

% methods that are benchmarked; cite papers that are being resembeled by each method
\begin{figure*}[t!]
\centering
\includegraphics[width=0.95\textwidth]{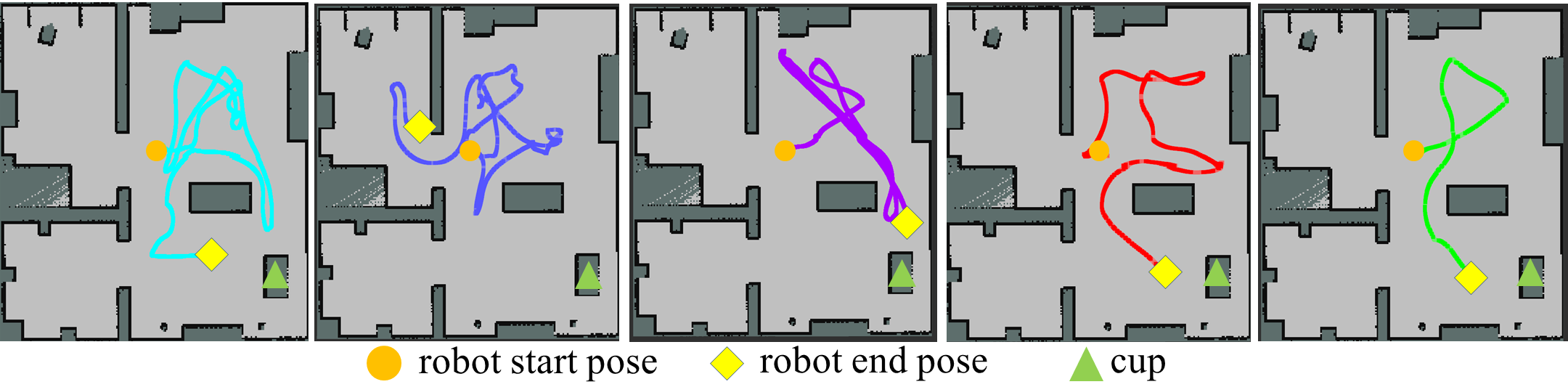}
\caption{Examples of search paths generated by each method while searching for \textit{cup}. Methods from left to right: UDS, IDS-Known-Static, IDS-Known-Dynamic, IDS-Unknown, IHS-Unknown. (Best viewed in color).}
\label{fig:search_path}
\end{figure*}

\begin{table*}[t!]
\centering
\begin{small}
\begin{tabular}{llrrrrr}
Target Object                   & Metrics          &    UDS & IDS known, static & IDS known, dynamic &   IDS unknown &    IHS unknown \\ \hline
\multirow{4}{*}{Coffee Machine} & Views            &   7.83 &              6.17 &               4.67 &          6.33 &  \textbf{3.67} \\
                                & Search Time (s)  &   107 &               76 &                60 &           75 &   \textbf{50} \\
                                & Search Path (m)  &  8.68 &             6.70 &              5.80 &         6.74 &  \textbf{4.93} \\
                                & Success Rate     & \textbf{1.0} &   \textbf{1.0} &           \textbf{1.0} &    \textbf{1.0} &   \textbf{1.0} \\ \hline
\multirow{4}{*}{Laptop}         & Views            &   11.00 &              12.50 &               7.17 &          5.67 &  \textbf{4.17} \\
                                & Search Time (s)  &   197 &              222 &               124 &           91 &   \textbf{78} \\
                                & Search Path (m)  & 28.27 &            26.86 &             13.13 &\textbf{7.69} &          8.40 \\
                                & Success Rate     &   0.83 &              0.50 &               1.00 &          1.00 &          1.00  \\ \hline
\multirow{4}{*}{Cup}            & Views            &  13.17 &          14.50 &                   12.67 &         11.83 &           \textbf{9.00} \\
                                & Search Time (s)  &   184 &         229 &                    189 &          185 &           \textbf{139} \\
                                & Search Path (m)  & 22.64 &          29.81 &             23.40 &        19.68 &         \textbf{13.91} \\
                                & Success Rate     &   0.83 &              0.33 &               0.83 &          0.83 &  \textbf{1.00}  \\
\end{tabular}
\end{small}
\caption{Benchmark results for object search in simulation experiments. Among methods that reached 100\% success rate, IHS unknown successfully found target objects within the smallest number of views and least search time.}
\label{tab:results}
\end{table*}

We set up an object detector in simulation that returns a detection of an object, if the object is in view, not fully occluded, and within the effective observation range. For large objects (e.g. sofa, bed, fridge), mid-sized objects (e.g. desk, table, sink), and small objects (e.g. cup, laptop, coffee machine), we assume an effective observation range of $5$m, $4$m, $2.5$m respectively.

We benchmark following methods:
%the baseline methods and our proposed method in searching for: a coffee machine, a laptop and a cup. The methods are explained as below:
\begin{itemize}
    \item \textbf{UDS}: Uninformed direct search (Eq.\ref{eq:ds}). The robot does not account for the spatial relations between the target and landmark objects (omitting Eq. \ref{eq:context_potential} in \slim). %The search strategy follows Eq.\ref{eq:ds}. 
    This baseline represents a naive approach for object search.%thus not considering the \textit{Context Potential} (Eq. \ref{eq:context_potential}).
    \item \textbf{IDS-Known-Static}: Informed direct search (Eq.\ref{eq:ds}) with a known prior on landmark object locations. The robot assumes that landmark objects are static at the locations provided by the prior. %The robot only maintains the belief over the target object's location, and accounts for the probabilistic spatial relations between the target object and landmarks. The search strategy follows Eq.\ref{eq:ds}. 
    This method resembles previous works~\cite{kollar2009utilizing, kunze2014using, toris2017temporal}.
    \item \textbf{IDS-Known-Dynamic}: Informed direct search (Eq.\ref{eq:ds}) with a known prior on landmark object locations. This is similar to IDS-Known-Static except that the robot does not assume the landmark objects to remain at the locations expressed in the prior.%, thus the robot maintains the belief over the target as well as the landmark objects. The particles for landmark objects are initialized around the locations given by the prior.
    \item \textbf{IDS-Unknown}: Informed direct search (Eq.\ref{eq:ds}) without prior on landmark object locations. The particles for landmark objects are initialized uniformly across the environment. %The search strategy follows Eq.\ref{eq:ds}. 
    This method resembles previous works~\cite{loncomilla2018bayesian, aydemir2010object}. %The particles for landmark objects are probabilistically initialized in different rooms based on the marginal belief of the spatial relation between the landmark object and each room.
    \item \textbf{IHS-Unknown}: Informed hybrid search (Eq.\ref{eq:hs}) without prior on landmark object locations.
\end{itemize}

All methods except for UDS are using the full \slim \ model. We assume that an occupancy-grid map of the environment is given. We also assume that the room types are accurately recognized across the environment. IDS-Known-$^\ast$ methods are provided with a noisy prior on landmark object locations which differ from the actual locations, to emulate the common cases where perfect knowledge about landmark locations is not available. %Specifically, the noisy prior on landmark object locations is generated by sampling a location from a Gaussian distribution centered at the ground truth location, with a variance of $2$m in both $x$ and $y$ direction. 
For all methods, the particles for the target object are initialized uniformly across the environment.

%We benchmark the object search performance in 2 different test scenarios, as shown in Figure \ref{fig:env_setup}. The target objects are placed at different locations in each test scenarios. 
For each target object, we run $6$ trials per method. In each trial, the robot starts at the same location, depicted in Figure \ref{fig:env_setup}. The object search is terminated if (1) the belief over the target object location has converged, or (2) the maximum search time of $5$mins has been exceeded. A trial is successful if the robot finds the target object before timeout. For each target object and each method, we measure the number of view poses, search time, distance travelled by the robot, and search success rate averaged across all trials.

The benchmark result is as shown in Table \ref{tab:results}. Examples of the resulting search path from each method are depicted in Figure \ref{fig:search_path}. As we can see, UDS is not as efficient because it is not making use of the spatial relations between the target and landmark objects in the environment. Given a noisy prior on landmark object locations, IDS-Known-Dynamic outperforms IDS-Known-Static because it accounts for the uncertainty of the landmark object locations, whereas IDS-Known-Static is misled by the noisy prior.

Given no prior information, IHS-unknown outperforms IDS-unknown because it encourages the robot to explore promising regions that contain the target and/or useful landmark objects, whereas IDS-unknown only considers promising regions that contain the target object. With IHS-unknown, the robot benefits from finding landmark objects which help narrow down the search region for the target object.

% The environment models a spacious, one-bedroom apartment consisting of three rooms which are split into four areas. Figure \ref{fig:env_setup} depicts the layout of the apartment with a combined bedroom and office space in the lower left, a kitchen area in the upper left, a living room area in the upper right and a bathroom area in the lower right. In this environment we have a simulated \textit{fetch} robot look for a \emph{cup}, \emph{coffee machine}, or \emph{laptop}. Each of these items can be found in one of two places, as shown by the latter two images in figure \ref{fig:env_setup}. We provide our system with a map of the environment and knowledge about where the kitchen, bedroom, bathroom and living room are to be found on the map. Additionally, we provide probabilities for objects occurring in a particular QSR as common sense knowledge. The specifics can be read from figure \ref{fig:common_sense_knowledge}

% For each of these methods we perform two searches for each object per possible location, yielding a total of 20 trials \todo{That seems wrong.}. In each trial, the robot has a total of five minutes to locate the object. If it fails to do so in this time, the attempt is viewed as failed.
% Aside from the failure rate, we measure the time taken and distance traveled, and count the number of view poses the robot examined before finding the object.
% We report the averages of these metrics in table \ref{tab:results}.

\subsubsection{Real-World Experiments:}
The real-world experiment is executed in an environment ($8$mx$8$m) that consists of a kitchen and a living room. The robot stays localized in the pre-mapped environment based on its LIDAR, and navigates based on a MPEPC based path planner~\cite{park2012IROS}. The target object is a cup, and landmark objects include table, sofa, coffee machine and sink. IHS-Unknown reached average success rate of 0.7 (7 out of 10 trials). The average number of view poses, search time and search path is $4.86$, $103$s, and $8.32$m repectively. The failure cases were due to false negative detection of the cup due to lighting (we used Faster R-CNN~\cite{ren2017faster} trained on COCO dataset~\cite{coco}). Examples of real-world experiments with a Fetch robot is available in online video \url{https://youtu.be/uWWJ5aV6ScE}.
\section{Conclusion}
In this paper we present an efficient active visual object search approach through the introduction of the \slim \ model. \slim \  simultaneously maintains the belief over target and landmark objects locations, while accounting for the probabilistic inter-object spatial relations. Further, we propose a hybrid search strategy that draws insights from both direct and indirect object search. Given noisy or no prior on landmark objects locations, we demonstrate the benefit of modeling landmark objects locations under uncertainty in \slim, and the hybrid search strategy that encourages the robot to explore promising areas that can contain the target and/or landmark objects in both simulation and real-world experiments.

%When noisy prior on landmark objects is given, we demonstrate the benefit of modeling landmark objects locations under uncertainty in \slim. When no prior on landmark objects is given, we demonstrate that the proposed hybrid search strategy outperforms a direct search strategy by encouraging the robot to explore areas that are promising and contain not only the target object but also landmark objects. We also show the proposed object search approach operating in real-world experiments.
%\balance
\bibliographystyle{abbrv}
\bibliography{main}

\end{document}